\documentclass{article} %
\usepackage{iclr2027_conference,times}

\usepackage{amsmath,amsfonts,bm}

\def\eqref#1{equation~\ref{#1}}

\def\1{\bm{1}}

\DeclareMathAlphabet{\mathsfit}{\encodingdefault}{\sfdefault}{m}{sl}
\SetMathAlphabet{\mathsfit}{bold}{\encodingdefault}{\sfdefault}{bx}{n}

\usepackage[utf8]{inputenc} %
\usepackage[T1]{fontenc}    %
\usepackage{hyperref}       %
\usepackage{url}            %
\usepackage{booktabs}       %
\usepackage{arydshln}      %
\usepackage{amsfonts}       %
\usepackage{nicefrac}       %
\usepackage{microtype}      %
\usepackage{xcolor}
\colorlet{blue}{black}
\hypersetup{hidelinks}         %

\usepackage{subcaption}
\usepackage{multirow}
\usepackage{graphicx}
\usepackage{amsmath}
\usepackage[most]{tcolorbox}
\usepackage{cleveref}
\usepackage{placeins}
\tcbset{
trajbox/.style={
  breakable,
  colback=gray!5,
  colframe=gray!40!black,
  boxrule=0.5pt,
  arc=3pt,
  left=4pt,
  right=4pt,
  top=4pt,
  bottom=4pt
}
}

\definecolor{qualSuccess}{HTML}{2E6B57}
\definecolor{qualFailure}{HTML}{A5453E}
\newcommand{\qualoutcome}[2]{\tcbox[on line,boxrule=0.4pt,arc=1pt,boxsep=0pt,left=3pt,right=3pt,top=1.2pt,bottom=1.2pt,colback=#1!8!white,colframe=#1,coltext=#1,fontupper=\footnotesize\bfseries]{#2}}

\title{Co-Evolving Agents: Learning from Failures as Hard Negatives}

\author{\normalfont
\begin{minipage}{\dimexpr\textwidth-2\tabcolsep\relax}
\centering
Yeonsung Jung$^{1}$, Trilok Padhi$^{2}$, Sina Shaham$^{3}$, Dipika Khullar$^{3}$,\\
Joonhyun Jeong$^{1,4}$, Ninareh Mehrabi$^{3,\dagger}$, Eunho Yang$^{1,5,\dagger}$\\[0.6em]
{\small
$^{1}$Graduate School of AI, KAIST \quad $^{2}$Georgia State University\\
$^{3}$Independent Researcher\\
$^{4}$NAVER Cloud \quad $^{5}$AITRICS}
\end{minipage}}

\iclrfinalcopy %
\begin{document}
\raggedbottom
\maketitle
\lhead{Preprint}
\begingroup
\renewcommand{\thefootnote}{\fnsymbol{footnote}}
\footnotetext[2]{Co-corresponding authors.}
\endgroup

{
\begin{abstract}
Self-evolving agents improve their performance on long-horizon tasks by learning from their own interactions with an environment. {A common approach uses the resulting failed trajectories as negatives for preference training. However, collecting an agent's own failures does not ensure that they provide informative supervision for further improvement. Obvious failures may be easy to reject without learning to identify errors in more plausible attempts. These plausible but incorrect trajectories can serve as \emph{hard negatives}, helping agents learn to distinguish successful behavior from failures with less obvious errors. To enable agents to generate and learn from such informative negatives, we propose a co-evolving framework that couples a target agent's self-improvement with a failure agent trained to generate hard negatives. The failure agent learns exclusively from both agents' failed trajectories to generate higher-reward failures, which the target agent uses as negatives for preference training. As the target policy improves, its failures provide new training data for the failure agent. The updated failure agent, in turn, provides new negatives for target training, allowing negative generation to adapt alongside the target policy. {Across online shopping, scientific reasoning, and interactive SQL querying, our framework improves average task reward by 5.7 over the baseline.} These results demonstrate the value of learning to generate and use informative failures for agent self-improvement.} We will make our code publicly available.
\end{abstract}

\section{Introduction}
\label{sec:intro}
Self-evolving agents improve their performance on long-horizon tasks by learning from their own interactions with an environment. Agents built on large language models (LLMs) carry out tasks such as online shopping, scientific reasoning, and interactive SQL querying by taking actions and observing the results~\citep{webshop,scienceworld,intercode}. {Completing these tasks requires a sequence of decisions, with each action informed by observations from earlier steps.} Starting from a policy trained on expert demonstrations~\citep{sft1,agenttuning}, they collect trajectories of these interactions and use them to update the policy~\citep{eto,agentR,qi2025webrl}. Repeating this process reduces the need for additional human supervision, which can be costly and time-consuming. Continued improvement depends on whether the self-generated trajectories provide useful supervision for the next round of training.

Recent studies use failed trajectories from these interactions as additional training data. Some construct trajectories that show how an agent can recover from an error and complete the task, then use them for iterative self-training~\citep{agentR}. Others train the agent to prefer expert demonstrations over its own failures~\citep{eto}. {Using the agent's own failures allows preference training to address mistakes made by its current policy. However, these negatives are collected from a policy trained to solve tasks, rather than one trained to generate informative failures. Some failures contain obvious errors and are easy to distinguish from expert demonstrations. Learning to reject them may provide limited supervision for recognizing errors in more plausible attempts. Additional sampling increases the number of candidates, but does not directly train the policy to generate informative negatives.}

Prior work in contrastive learning and preference optimization has demonstrated the benefits of selecting \emph{hard negatives} that are difficult to distinguish from positive or preferred examples~\citep{hardnegative,chen2026mcpo}. {For agent training, this motivates using failures that require a closer comparison with successful behavior to identify what went wrong.} Consider a shopping task that specifies a product type, color, and budget. Selecting an unrelated product is an obvious failure. Selecting the requested type and color while exceeding the budget satisfies part of the instruction but still fails the task. Comparing this attempt with a successful purchase highlights the missed budget constraint. {As the target policy improves, failures that were once difficult to distinguish may become easy to reject. Repeated training on these familiar failure patterns risks overfitting to the observed negatives rather than learning to recognize errors in new trajectories. This motivates training a negative generator on the changing experience of the target policy.}

{To integrate negative generation with agent self-improvement, we propose a co-evolving framework that trains a failure agent to generate hard negatives and a target agent to learn from them (Figure~\ref{fig:main}).} The failure agent learns exclusively from both agents' failed trajectories, with the objective of generating high-reward failures. We use environment reward, which reflects partial task completion, as a proxy for negative difficulty. {The target agent learns to prefer expert demonstrations over failures generated by either agent. Training a separate failure agent therefore adds a source of negatives whose generation is explicitly optimized for target training. This makes learning to generate hard negatives part of the agent's self-improvement process.}

{The two agents co-evolve through an exchange of self-generated trajectories. After each round of training, the target agent collects new trajectories with its updated policy. Its failures provide new training data for the failure agent. In turn, trajectories from the updated failure policy provide negatives for subsequent target training. Thus, target training changes the experience available to the failure agent, while failure-agent training changes the negatives available to the target agent.} Preference labels come from existing demonstrations and environment rewards, so no additional human preference annotations are required. The failure agent adds trajectory collection and optimization costs during training; only the target agent is used at inference time.

{We evaluate our framework on WebShop for online shopping~\citep{webshop}, ScienceWorld for scientific reasoning~\citep{scienceworld}, and InterCodeSQL for interactive SQL querying~\citep{intercode}. Across these tasks, our framework consistently achieves higher task rewards than the baseline, with an average improvement of $5.7$. These improvements suggest that learning to generate hard negatives provides useful supervision beyond the target agent's own failures and supports agent self-improvement across domains.}

Our contributions are threefold:
\begin{itemize}
    \item {We introduce a failure agent that learns from failed trajectories to generate high-reward failures as hard negatives.}
    \item {We develop a co-evolving framework that uses target-agent failures to train the failure agent and the resulting hard negatives to improve the target agent.}
    \item {We evaluate our framework across three task domains and multiple base models, demonstrating performance gains from incorporating learned hard-negative generation into agent self-improvement.}
\end{itemize}

\section{Related Work}
\paragraph{Self-evolving agents.}
Self-evolving agents improve by training on trajectories collected through their own interactions. A common pipeline first fine-tunes the agent on demonstrations~\citep{sft1,agenttuning} and then updates it using successful or failed rollouts~\citep{rft,eto,agentR}. Existing methods retain successful rollouts, pair failures with demonstrations, assign feedback to individual steps, or train on trajectories that show how to recover from errors~\citep{rft,eto,IPR,agentR}. Online curricula can further adjust the training tasks as the agent improves~\citep{qi2025webrl}. These methods use self-generated trajectories to improve the agent or adapt its training curriculum. However, failures collected from the current agent can become less informative after it learns to avoid earlier error patterns.

\paragraph{Learning from generated preferences.}
Preference optimization trains a model to prefer one output over another without learning a separate reward model~\citep{dpo}. Recent methods construct preferences from model-generated evaluations, combine feedback at multiple levels of a trajectory, or allow models with different capabilities to learn from one another~\citep{yuan2024selfrewarding,gao2026hpl,macpo}. Negative agents have also been trained on dispreferred responses to supply negatives for alignment; these generators remain fixed after initialization~\citep{macpo}.

\begin{figure}[!t]
\centering
\includegraphics[width=\linewidth]{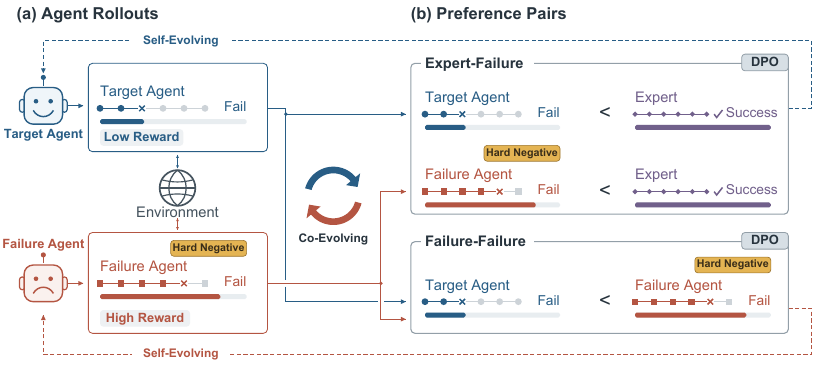}
\caption{{Overview of our co-evolving framework. Both agents generate trajectories for repeated training. The target agent prefers expert demonstrations over failures; the failure agent prefers higher-reward failures over lower-reward ones, regardless of source. The highlighted hard negative is rejected during target agent training but preferred during failure agent training. Colors denote trajectory sources, bars show rewards, and dashed loops indicate repeated updates.}}
\label{fig:main}
\end{figure}

\paragraph{Hard negatives and sampling.}
Hard negatives are plausible failures that are more difficult to distinguish from successful behavior than obvious failures. Prior work shows that such negatives can provide stronger contrastive supervision~\citep{hardnegative} and proposes sampling rejected responses that are difficult for the current policy~\citep{chen2026mcpo}. Other work allocates more trials to difficult prompts to collect successful solutions~\citep{dart}. Most focus on response-level preference learning or on collecting successful solutions, while hard-negative generation for failed multi-turn interactions remains less explored.

\section{Preliminaries}
\label{sec:preliminaries}
{We model agent--environment interaction as a partially observable Markov decision process, following \citet{eto}. Let $\mathcal{U}$, $\mathcal{S}$, $\mathcal{A}$, and $\mathcal{O}$ denote the instruction, state, action, and observation spaces, respectively. The transition model $T(s'\mid s,a)$ describes how an action changes the latent state, and the observation model $Z(o\mid s',a)$ describes the feedback returned to the agent. Instructions, actions, and observations are represented as text.}

{Given an instruction $u\in\mathcal{U}$, the agent selects each action from its interaction history:}
\[
h_j=(u,a_1,o_1,\ldots,a_{j-1},o_{j-1}),
\qquad a_j\sim\pi_\theta(\cdot\mid h_j).
\]
{For the first action, $h_1=u$, where $u$ includes any initial environment observation. Interaction continues until the episode terminates or the step budget is reached. A trajectory $\tau=(u,a_1,o_1,\ldots,a_n,o_n)$ includes the observation returned after the final action. We write $\pi_\theta(\tau\mid u)$ for the likelihood of the agent's actions conditioned on the recorded observations:}
{
\begin{equation}
\log\pi_\theta(\tau\mid u)=\sum_{j=1}^{n}\log\pi_\theta(a_j\mid h_j).
\label{eq:trajectory_likelihood}
\end{equation}
}
{Each action likelihood is evaluated autoregressively over its tokens. Instruction and observation tokens provide context but do not contribute to the training loss.}

{The environment assigns an episode-level reward $r(u,\tau)\in[0,1]$ to the trajectory. This reward measures task completion and need not be a sum of step rewards. Each environment specifies a success condition for the task.}

\section{Method}
\label{sec:method}
{Our framework trains a target agent to solve tasks and a failure agent to generate hard negatives. The target agent learns to prefer expert demonstrations over failed trajectories. The failure agent learns to generate high-reward failures by comparing unsuccessful trajectories from both agents. Both agents start from the same supervised checkpoint and exchange self-generated trajectories over successive training iterations (Figure~\ref{fig:main}).}

\subsection{Supervised Initialization}
\label{subsec:bc}
{Supervised fine-tuning on expert demonstrations is a common initialization step before further agent training~\citep{eto,agenttuning}. We use a small set of domain-specific expert demonstrations $\mathcal{D}_{\mathrm{exp}}$ to train an initial policy for the task environment:}
{
\begin{equation}
\mathcal{L}_{\mathrm{SFT}}(\theta)
=-\mathbb{E}_{(u,\tau_{\mathrm{exp}})\sim\mathcal{D}_{\mathrm{exp}}}
\left[\log\pi_\theta(\tau_{\mathrm{exp}}\mid u)\right].
\label{eq:sft}
\end{equation}
}
{The resulting checkpoint initializes both the target agent $\pi_\theta$ and the failure agent $\pi_\phi$, giving them the same starting point before they learn from different preference data. Demonstration sources are described in Appendix~\ref{appendix:datasets}.}

\subsection{Learning to Generate Hard Negatives}
\label{subsec:failure-agent}
{We train the failure agent to generate high-reward failures that serve as hard negatives for training the target agent. Environment reward measures partial task completion, which we use as a proxy for negative difficulty. This proxy does not directly measure how difficult a failure is for the target agent to distinguish from success.}

{We obtain the training signal by comparing failures from the two agents on the same instruction. Let $\mathcal{F}_\theta(u)$ and $\mathcal{F}_\phi(u)$ denote their respective sets of failed trajectories for instruction $u$. For each pair $\tau_\theta\in\mathcal{F}_\theta(u)$ and $\tau_\phi\in\mathcal{F}_\phi(u)$, we prefer the trajectory with higher reward:}
{
\begin{equation}
(\tau^+,\tau^-)=
\begin{cases}
(\tau_\theta,\tau_\phi), & r(u,\tau_\theta)>r(u,\tau_\phi),\\
(\tau_\phi,\tau_\theta), & r(u,\tau_\phi)>r(u,\tau_\theta).
\end{cases}
\label{eq:failure_pairs}
\end{equation}
}
{We discard ties and collect preference triplets $(u,\tau^+,\tau^-)$ in $\mathcal{D}_{\mathrm{fail}}$. Both $\tau^+$ and $\tau^-$ are failed trajectories, with $r(u,\tau^+)>r(u,\tau^-)$. The preferred trajectory can come from either agent.}

{We optimize the failure agent on $\mathcal{D}_{\mathrm{fail}}$ using Direct Preference Optimization (DPO)~\citep{dpo}. For a policy $\pi$ and a reference policy $\pi_{\mathrm{ref}}$, the loss for a preference pair is}
{
\begin{equation}
\ell_{\mathrm{DPO}}(\pi;u,\tau^+,\tau^-)
=-\log\sigma\!\left[\beta\!\left(
\log\frac{\pi(\tau^+\mid u)}{\pi_{\mathrm{ref}}(\tau^+\mid u)}
-\log\frac{\pi(\tau^-\mid u)}{\pi_{\mathrm{ref}}(\tau^-\mid u)}
\right)\right].
\label{eq:dpo}
\end{equation}
}
{Here, $\sigma$ is the logistic sigmoid and $\beta>0$ is a scaling parameter. Each agent uses its own reference policy, held fixed during each optimization phase. We train the failure agent by minimizing}
{
\begin{equation}
\mathcal{L}_{\mathrm{fail}}(\phi)
=\mathbb{E}_{(u,\tau^+,\tau^-)\sim\mathcal{D}_{\mathrm{fail}}}
\left[\ell_{\mathrm{DPO}}(\pi_\phi;u,\tau^+,\tau^-)\right].
\label{eq:failure_objective}
\end{equation}
}

{The failure agent is trained to prefer higher-reward failures using only failure--failure pairs. After the shared supervised initialization, its updates use no expert demonstrations or auxiliary SFT loss. The failure agent may still generate successful trajectories, which are excluded from its training data.}

\begin{table}[!t]
\color{black}
\centering
\caption{Benchmark statistics. ScienceWorld includes seen and unseen evaluation splits. Max Turns denotes the interaction limit per episode.}
\label{tab:overview}
\setlength{\tabcolsep}{4pt}
\begin{tabular}{lccccc}
\toprule
\multirow{2}{*}{Dataset} &
\multirow{2}{*}{Train} &
\multicolumn{2}{c}{Test} &
\multirow{2}{*}{Action Space} &
\multirow{2}{*}{Max Turns} \\
\cmidrule(lr){3-4}
 &  & Seen & Unseen &  &  \\
\midrule
WebShop        & 1,824 & 200 & --  & 8             & 10 \\
ScienceWorld   & 1,483 & 194 & 211 & 16            & 100 \\
InterCodeSQL   & 1,500 & 200 & --  & $\infty$ & 10 \\
\bottomrule
\end{tabular}

\end{table}

\subsection{Learning from Hard Negatives}
\label{subsec:co-evolution}
{We augment the target agent's own failures with hard negatives generated by the failure agent. For each instruction, we pair each collected failure with an expert demonstration, using the demonstration as the preferred trajectory and the failure as the rejected trajectory. This gives the target preference dataset}
{
\begin{equation}
\mathcal{D}_{\mathrm{tgt}}=\big\{(u,\tau_{\mathrm{exp}},\tau^-): (u,\tau_{\mathrm{exp}})\in\mathcal{D}_{\mathrm{exp}},\;\tau^-\in\mathcal{F}_\theta(u)\cup\mathcal{F}_\phi(u)\big\}.
\label{eq:target_pairs}
\end{equation}
}

{We apply the DPO loss in Equation~\ref{eq:dpo} to these pairs. DPO learns relative preferences, but its updates can become dominated by suppressing rejected responses without sufficiently reinforcing preferred responses~\citep{yan2025dpo3d}. To stabilize target training, we add an auxiliary SFT loss on the expert demonstrations. This term directly increases the likelihood of expert actions, complementing the relative preference signal from DPO.}

{We normalize pair weights within each instruction so that instructions with more failures do not dominate training. Let $\mathbb{E}_w$ denote the weighted average over $\mathcal{D}_{\mathrm{tgt}}$, with equal total weight for each instruction (Appendix~\ref{appendix:weighting}). The combined objective is}
{
\begin{equation}
\mathcal{L}_{\mathrm{tgt}}(\theta)=\mathbb{E}_{w}\left[\ell_{\mathrm{DPO}}(\pi_\theta;u,\tau_{\mathrm{exp}},\tau^-)\right]+\lambda_{\mathrm{SFT}}\,\mathbb{E}_{w}\left[-\log\pi_\theta(\tau_{\mathrm{exp}}\mid u)\right].
\label{eq:target_objective}
\end{equation}
}
{Here, $\lambda_{\mathrm{SFT}}$ controls the weight of the auxiliary SFT loss. By incorporating failures from a separately trained failure agent, target training extends beyond the target agent's own mistakes, which may help it generalize beyond the failure patterns encountered in its own rollouts.}

\suppressfloats[t]
\begin{table}[t]
\color{black}
\centering
\caption{Main results on WebShop, ScienceWorld, and InterCodeSQL across three base models. Results are average task rewards on a $0$--$100$ scale.}
\label{tab:main_combined}
\label{tab:main_qwen3_4b}
\small
\setlength{\tabcolsep}{3pt}
\begin{tabular}{@{}llccccc@{}}
\toprule
\multirow{2}{*}{\textbf{Model}} & \multirow{2}{*}{\textbf{Method}} & \multirow{2}{*}{\textbf{WebShop}} & \multicolumn{2}{c}{\textbf{ScienceWorld}} & \multirow{2}{*}{\textbf{InterCodeSQL}} & \multirow{2}{*}{\textbf{Avg.}} \\
\cmidrule(lr){4-5}
 & & & \textbf{Seen} & \textbf{Unseen} & & \\
\midrule
\multirow{3}{*}{Qwen3-4B} & SFT & 66.00 & 83.30 & 67.90 & 58.10 & 68.83 \\
 & ETO & 72.90 & 80.70 & 83.10 & 60.70 & 74.35 \\
 & Ours & \textbf{74.10} & \textbf{87.60} & \textbf{88.30} & \textbf{63.00} & \textbf{78.25} \\
\midrule
\multirow{5}{*}{Qwen3.5-9B} & SFT & 55.15 & 79.24 & 73.80 & 68.59 & 69.20 \\
 & Agent-R & 55.24 & 69.68 & 56.66 & 65.31 & 61.72 \\
 & RFT & 56.41 & 80.00 & 74.57 & 69.57 & 70.14 \\
 & ETO & 68.06 & 80.47 & 75.40 & 68.94 & 73.22 \\
 & Ours & \textbf{71.30} & \textbf{88.60} & \textbf{84.72} & \textbf{70.63} & \textbf{78.81} \\
\midrule
\multirow{5}{*}{Gemma-4-12B-it} & SFT & 71.10 & 70.71 & 64.96 & 66.77 & 68.39 \\
 & Agent-R & 69.70 & 70.21 & 65.19 & 67.32 & 68.10 \\
 & RFT & 72.22 & 77.25 & 73.29 & 70.82 & 73.40 \\
 & ETO & 71.67 & 77.11 & 69.59 & 70.36 & 72.18 \\
 & Ours & \textbf{73.49} & \textbf{87.55} & \textbf{84.42} & \textbf{73.38} & \textbf{79.71} \\
\bottomrule
\end{tabular}
\end{table}

\subsection{Co-Evolution through Trajectory Exchange}
\label{subsec:training_loop}
{We train the two agents iteratively, starting from the same supervised checkpoint. In each iteration, both agents collect trajectories through environment interaction. We construct $\mathcal{D}_{\mathrm{fail}}$ and $\mathcal{D}_{\mathrm{tgt}}$ from the collected failures and expert demonstrations, as described in Sections~\ref{subsec:failure-agent} and~\ref{subsec:co-evolution}. Each agent is then updated with its respective objective in Equations~\ref{eq:failure_objective} and~\ref{eq:target_objective}. We repeat this procedure using trajectories generated by the updated agents.}

{The failure agent learns from the target agent's current failures, so its training data change as the target agent is updated. In turn, the updated failure agent generates new negatives for subsequent target training. This feedback loop allows hard-negative generation to adapt as the target agent learns. Both agents therefore learn from their own interactions with the environment and from trajectories generated by the other agent.}

{The failure agent supports the target agent during training by generating hard negatives for preference learning. At inference time, we use only the target agent to interact with the environment and complete tasks.}

\section{Experiments}
\label{sec:experiments}
We evaluate whether learning to generate hard negatives improves agent self-improvement across three task domains and multiple base models. Performance comparisons assess the full framework, trajectory analyses examine the generated training data, and ablations test the roles of the failure agent, continued co-evolution, and auxiliary expert supervision.

\subsection{Experimental Setup}
\label{sec:setup}
We consider a setting with a small set of expert demonstrations for each environment.

\paragraph{Environments.}
{We evaluate on three interactive environments. WebShop requires agents to search for products and purchase an item that satisfies the user's requirements~\citep{webshop}. ScienceWorld requires agents to carry out scientific procedures that combine navigation, object manipulation, and reasoning across multiple steps~\citep{scienceworld}. InterCodeSQL requires agents to answer database questions by generating SQL queries and refining them using execution feedback~\citep{intercode}. Trajectories span up to 10 turns in WebShop and InterCodeSQL and up to 100 turns in ScienceWorld.} All three environments return rewards in $[0,1]$ and provide partial credit for task progress. We report average rewards on a $0$--$100$ scale. ScienceWorld includes separate seen and unseen evaluation splits. Table~\ref{tab:overview} summarizes dataset sizes and interaction budgets, and Appendix~\ref{appendix:datasets} describes the reward rules and demonstration sources.

\paragraph{Models.}
We evaluate Qwen3-4B-Instruct-2507~\citep{qwen3}, Qwen3.5-9B~\citep{qwen35}, and Gemma-4-12B-it~\citep{gemma4}. To assess whether the gains extend to earlier model generations, we also evaluate Llama-2-7B-Chat and Llama-2-13B-Chat~\citep{llama2}, with results reported in Appendix~\ref{appendix:additional_results}. The target and failure agents use the same model architecture and start from the same supervised checkpoint.

\paragraph{Baselines.}
SFT is the supervised starting policy. RFT trains on successful generated trajectories~\citep{rft}, ETO uses expert--failure preferences~\citep{eto}, and Agent-R uses iterative self-training for error recovery~\citep{agentR}. Self-Rewarding Language Models (SRLM) construct preferences through self-evaluation of generated candidates~\citep{yuan2024selfrewarding}. To compare learned hard-negative generation with additional sampling, Best-of-$N$ samples four trajectories from the target agent for each instruction and selects the one with the highest environment reward for preference construction. PPO~\citep{ppo} and prompt-only GPT baselines appear in the additional model comparisons in Appendix~\ref{appendix:additional_results}.

\paragraph{Implementation details.}
We initialize both agents with SFT on expert demonstrations and use AdamW~\citep{adamw} with three epochs per optimization phase. The default target objective uses $\lambda_{\mathrm{SFT}}=0.1$; Table~\ref{tab:lambda_ablation} varies this weight. Multiple-candidate comparisons use $N=4$. Rollout counts are defined per instruction per generating agent and do not imply matched total training costs. Evaluation uses the target agent alone. Avg. is the arithmetic mean of the four evaluation columns. Further optimization settings and computational resources are given in Appendix~\ref{appendix:Implementation}.

\FloatBarrier
\subsection{Main Results}
\label{sec:main_results}
{Table~\ref{tab:main_combined} reports task performance across three environments and three base models. ETO trains the target agent on expert--failure pairs using its own failed trajectories. The comparison with ETO evaluates the benefit of learning to generate hard negatives with a separate failure agent.}

{Our framework achieves the highest task reward across the evaluated models and environments, with an average improvement of $5.7$ over ETO. Performance improves across all three environments, especially on ScienceWorld, the most challenging long-horizon environment in our evaluation. Its tasks combine scientific reasoning with multiple dependent actions over trajectories of up to 100 turns. These results suggest that a trained failure agent provides useful supervision beyond the target agent's own failures, supporting agent self-improvement across models and tasks.}

\FloatBarrier
\subsection{Comparison with Additional Sampling}
\label{sec:multiple_rollouts}
To evaluate whether additional sampling can match the benefits of learned hard-negative generation, we compare our framework with SRLM and Best-of-$N$ under the ETO training setup. Both methods sample four target trajectories per instruction. SRLM assigns rewards through self-evaluation and uses them to construct preference pairs without environment rewards. Best-of-$N$ instead uses environment rewards to select the highest-reward trajectory for preference construction.

{Our framework achieves the highest task rewards across both models and all environments (Table~\ref{tab:multiple_rollouts}). SRLM constructs pairs using self-assigned rewards and generally underperforms methods using environment rewards. Best-of-$N$ selects the highest-reward rollouts, but those matching the expert reward yield no valid expert--failure pairs and can reduce opportunities to learn from failures. These results support training a failure agent to generate informative negatives for self-improvement.}

\begin{table}[!t]
\centering
\caption{Comparison with methods that use multiple rollouts during training. SRLM and Best-of-$N$ use $N=4$; ETO and Ours use one rollout per instruction per generating agent.}
\label{tab:multiple_rollouts}
\small
\setlength{\tabcolsep}{4pt}
\renewcommand{\arraystretch}{1.03}
\begin{tabular}{@{}llccccc@{}}
\toprule
\multirow{2}{*}{Model} & \multirow{2}{*}{Method} & \multirow{2}{*}{WebShop} & \multicolumn{2}{c}{ScienceWorld} & \multirow{2}{*}{InterCodeSQL} & \multirow{2}{*}{Avg.} \\
\cmidrule(lr){4-5}
 & & & Seen & Unseen & & \\
\midrule
\multirow{4}{*}{Qwen3.5-9B} & ETO & 68.06 & 80.47 & 75.40 & 68.94 & 73.22 \\
 & SRLM & 54.09 & 78.11 & 74.09 & 66.34 & 68.16 \\
 & Best-of-$N$ & 71.14 & 82.87 & 75.94 & 65.02 & {73.74} \\
 & Ours & \textbf{71.30} & \textbf{88.60} & \textbf{84.72} & \textbf{70.63} & \textbf{78.81} \\
\midrule
\multirow{4}{*}{Gemma-4-12B-it} & ETO & 71.67 & 77.11 & 69.59 & 70.36 & 72.18 \\
 & SRLM & 72.27 & 72.93 & 68.52 & 71.75 & {71.37} \\
 & Best-of-$N$ & 71.44 & 80.72 & 72.64 & 71.05 & 73.96 \\
 & Ours & \textbf{73.49} & \textbf{87.55} & \textbf{84.42} & \textbf{73.38} & \textbf{79.71} \\
\bottomrule
\end{tabular}
\end{table}
\FloatBarrier
\subsection{Qualitative Results}
\label{sec:qualitative}
\begingroup
We analyze matched ScienceWorld examples to compare target performance and the failures used for training. ETO learns from its own failures and is among the strongest baselines in Table~\ref{tab:main_combined}. Both comparisons below use Gemma-4-12B-it.

\begin{tcolorbox}[trajbox,before skip=5pt,after skip=5pt,boxsep=1pt,top=3pt,bottom=3pt,title={Target Performance: Pollination (ScienceWorld)},title after break={Target Performance: Pollination (ScienceWorld) (continued)}]
\textbf{Instruction:} Grow an orange by growing several plants and allowing cross-pollination. Focus on the grown orange to finish the task.\par\vspace{3pt}\hrule\vspace{3pt}
\textbf{ETO target} \hfill Reward 0.446 \quad \qualoutcome{qualFailure}{FAILURE}\par\nobreak
Plants three orange seeds and waits until the trees have flowers, but \textbf{leaves the bee hive closed}. It continues waiting and attempts to focus on an orange before any fruit appears, leaving the pollination step incomplete.
\par\vspace{3pt}\hrule\vspace{3pt}
\textbf{Ours target} \hfill Reward 1.000 \quad \qualoutcome{qualSuccess}{SUCCESS}\par\nobreak
Plants the seeds, waits for flowers, and \textbf{opens the bee hive}. Later observations show a bee and oranges on the trees; after resolving an ambiguous object selection, it \textbf{focuses on a grown orange and completes the task}.
\par\vspace{3pt}\hrule\vspace{3pt}
\textbf{Key difference:} Both agents grow flowering trees, but only Ours enables pollination and follows the resulting fruit through to task completion.
\end{tcolorbox}

\begin{tcolorbox}[trajbox,before skip=5pt,after skip=5pt,boxsep=1pt,top=3pt,bottom=3pt,title={Training Negatives: Measurement and Decision (ScienceWorld)},title after break={Training Negatives: Measurement and Decision (ScienceWorld) (continued)}]
\textbf{Instruction:} Measure the melting point of gallium. Focus on the orange box if it is above $200^\circ$C, or the yellow box if it is below $200^\circ$C.\par\vspace{3pt}\hrule\vspace{3pt}
\textbf{ETO negative} \hfill Reward 0.403 \quad \qualoutcome{qualFailure}{FAILURE}\par\nobreak
Retrieves and focuses on the thermometer, then attempts to focus on the gallium-containing jug. The environment \textbf{selects the jug itself}, and the rollout \textbf{ends before the agent measures the material} or selects an answer box.
\par\vspace{3pt}\hrule\vspace{3pt}
\textbf{Ours hard negative} \hfill Reward 0.747 \quad \qualoutcome{qualFailure}{FAILURE}\par\nobreak
Focuses on gallium, opens the freezer, cools the sample, and obtains a $7^\circ$C reading. It nevertheless \textbf{selects the orange box for the above-$200^\circ$C condition}; the measurement also \textbf{does not establish the melting point} requested by the instruction.
\par\vspace{3pt}\hrule\vspace{3pt}
\textbf{Key difference:} The hard negative completes more of the procedure but still reaches an unsupported conclusion. The error concerns what the measurement establishes and how it should guide the final decision.
\end{tcolorbox}

The target comparison highlights a missing dependency in a multi-step task: flowering must be followed by pollination to produce fruit. The training comparison shows a different distinction: completing several experimental steps does not ensure a valid conclusion. Such failures offer a concrete contrast with successful demonstrations beyond early procedural errors. Appendix~\ref{appendix:quality} provides three comparisons per environment, including the corresponding actions and observations.

\endgroup

\FloatBarrier
\subsection{Failure Reward Analysis}
\label{fail_analysis}
We examine whether the failure agent provides more high-reward negatives for target training than ETO. Figure~\ref{fig:iter1_failure_composition}(a) groups failures into Easy, Medium, and Hard bins, corresponding to the lower, middle, and upper thirds of the reward range, respectively. These bins are used only for analysis.

\begin{figure}[!t]
\centering
\includegraphics[width=\linewidth]{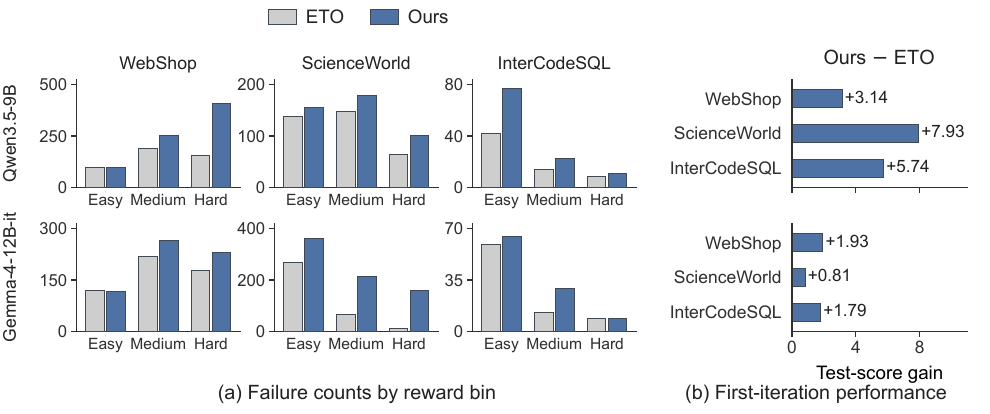}
\caption{\textcolor{black}{Failure rewards and early target performance. (a) Failure counts by reward bin for Qwen3.5-9B (top) and Gemma-4-12B-it (bottom). Each panel uses its own count scale. (b) Target performance gains over ETO at iteration 1. Reward bins are used only for analysis.}}
\label{fig:iter1_failure_composition}
\end{figure}

Our framework produces more Hard failures in most model--environment settings, providing a larger pool of high-reward negatives for target training. Their proportion increases on WebShop and ScienceWorld but decreases on InterCodeSQL, indicating that greater availability does not imply a uniform shift in the reward distribution. Panel (b) shows target performance gains over ETO across all evaluated settings. This analysis characterizes the training data but does not isolate their effect on performance.

\FloatBarrier

\subsection{Component Ablations}
\label{sec:ablations}
\label{sec:failure_agent_ablation}
\label{sec:coevolution_ablation}

\paragraph{Failure agent.}
{To assess the contribution of the failure agent, we replace it with a positive agent trained with the same objective as the target agent. We keep the total parameter count unchanged. This separate positive agent provides additional trajectories to augment the target agent's training data. As shown in Table~\ref{tab:agent_ablations}, training with a failure agent yields higher target agent reward than training with a positive agent across the reported settings, with the largest gains on ScienceWorld's multi-step scientific tasks. These results support training the additional agent specifically to generate hard negatives for the target agent.}

\paragraph{Co-evolution.}
{To evaluate the effect of co-evolution, we freeze the failure agent after training it in iteration 1. The failure agent has thus learned from the target agent's initial failures, but does not learn from the target agent's subsequent failures. The target agent continues learning from new negatives generated by this frozen failure agent. As shown in Table~\ref{tab:agent_ablations}, continuing to update the failure agent yields higher target agent reward than freezing it across all reported settings, with larger gains on ScienceWorld. These results suggest that learning from the target agent's recent failures helps the failure agent generate negatives that are more informative for the target agent at its current stage of learning.}

\begin{table}[!htb]
\centering
\caption{Ablations of the failure agent and co-evolution. w/o Failure Agent replaces the failure agent with a positive agent trained with the target objective; w/o Co-Evolution freezes the failure agent after iteration 1.}
\label{tab:agent_ablations}
\small
\setlength{\tabcolsep}{4pt}
\begin{tabular}{@{}llccccc@{}}
\toprule
\multirow{2}{*}{Model} & \multirow{2}{*}{Method} & \multirow{2}{*}{WebShop} & \multicolumn{2}{c}{ScienceWorld} & \multirow{2}{*}{InterCodeSQL} & \multirow{2}{*}{Avg.} \\
\cmidrule(lr){4-5}
 & & & Seen & Unseen & & \\
\midrule
\multirow{5}{*}{Qwen3.5-9B} & SFT & 55.15 & 79.24 & 73.80 & 68.59 & 69.20 \\
 & ETO & 68.06 & 80.47 & 75.40 & 68.94 & 73.22 \\
\cmidrule[0.3pt](l){2-7}
 & \textbf{Ours} & \textbf{71.30} & \textbf{88.60} & \textbf{84.72} & \textbf{70.63} & \textbf{78.81} \\
 & \hspace{0.8em}w/o Co-Evolution & 71.10 & 87.05 & 79.52 & 70.27 & 76.99 \\
 & \hspace{0.8em}w/o Failure Agent & 70.93 & 68.62 & 63.03 & 68.71 & 67.82 \\
\midrule
\multirow{5}{*}{Gemma-4-12B-it} & SFT & 71.10 & 70.71 & 64.96 & 66.77 & 68.39 \\
 & ETO & 71.67 & 77.11 & 69.59 & 70.36 & 72.18 \\
\cmidrule[0.3pt](l){2-7}
 & \textbf{Ours} & \textbf{73.49} & \textbf{87.55} & \textbf{84.42} & \textbf{73.38} & \textbf{79.71} \\
 & \hspace{0.8em}w/o Co-Evolution & 73.21 & 81.29 & 76.43 & 71.84 & 75.69 \\
 & \hspace{0.8em}w/o Failure Agent & 72.99 & 77.33 & 72.52 & 71.38 & 73.56 \\
\bottomrule
\end{tabular}
\end{table}

\FloatBarrier

\subsection{Sensitivity to the SFT Weight}
\label{sec:sft_weight}
\newsavebox{\sftweighttablebox}
\sbox{\sftweighttablebox}{%
\small
\setlength{\tabcolsep}{3pt}
\begin{tabular}{@{}llccc@{}}
\toprule
 & & \multicolumn{3}{c}{$\lambda_{\mathrm{SFT}}$} \\
\cmidrule(lr){3-5}
Model & Split & 0.01 & 0.1 & 0.5 \\
\midrule
\multirow{2}{*}{Qwen3.5} & Seen & 85.17 & \textbf{88.60} & 87.90 \\
 & Unseen & 80.56 & \textbf{84.72} & 76.41 \\
\midrule
\multirow{2}{*}{Gemma-4} & Seen & 84.84 & \textbf{87.55} & 87.05 \\
 & Unseen & 83.65 & \textbf{84.42} & 83.89 \\
\bottomrule
\end{tabular}%
}
\par
\noindent\begin{minipage}[t]{\dimexpr\linewidth-\wd\sftweighttablebox-10pt\relax}
\vspace{0pt}
The auxiliary SFT loss reinforces expert actions to stabilize DPO training. We use $\lambda_{\mathrm{SFT}}=0.1$ across all models and environments in the main experiments and vary this weight on ScienceWorld to assess sensitivity (Table~\ref{tab:lambda_ablation}). Both models score highest at the shared weight; increasing it does not improve performance. This supports using SFT as a moderately weighted auxiliary objective. Appendix~\ref{appendix:sft_weight} reports the Llama-2 results on WebShop.
\end{minipage}\hspace{10pt}%
\begin{minipage}[t]{\wd\sftweighttablebox}
\vspace{0pt}
\centering
\captionsetup{hypcap=false,width=\wd\sftweighttablebox}
\captionof{table}{SFT weight ablation.}
\label{tab:lambda_ablation}
\usebox{\sftweighttablebox}
\end{minipage}
\par

\FloatBarrier
\section{Conclusion}
{We introduced a co-evolving framework that learns to generate hard negatives for agent self-improvement. The failure agent learns from both agents' failed trajectories to generate higher-reward failures, which the target agent uses for preference training. As the target agent improves, its new failures guide further learning by the failure agent, allowing negative generation to adapt throughout training. Experiments on WebShop, ScienceWorld, and InterCodeSQL show improvements over baselines, especially on ScienceWorld's challenging long-horizon tasks. The failure agent yields higher target reward than an additional positive agent used to augment the training data. These findings highlight the value of learning to generate informative negatives for agent self-improvement.}

\paragraph{Limitations.}
{Training the failure agent increases rollout and optimization costs, although inference uses only the target agent. The failure agent relies on reward differences, which provide no preference signal for tied failures and do not directly measure negative difficulty. Future work could explore self-verification to identify informative failures without relying on environment rewards.}

\newpage
\subsection*{AI Use Statement}
We used LLMs or other AI tools to aid or polish writing, for research ideation or execution, to draft sections of the paper, and for the implementation of experimental code. The authors guided and reviewed all AI-assisted work and take full responsibility for the final manuscript and associated artifacts.

\subsection*{Reproducibility Statement}
Section~\ref{sec:method} describes trajectory construction, the construction of preference pairs, and the learning objectives for both agents. Section~\ref{sec:setup} specifies the benchmarks, evaluated models, baselines, and evaluation metrics. Appendix~\ref{appendix:datasets} documents the demonstration sources and reward rules, Appendix~\ref{appendix:Implementation} summarizes the optimization settings and computational resources, and Appendix~\ref{appendix:weighting} details the sample weights and supervision from expert demonstrations used to train the target agent. The accompanying implementation details distinguish the agents' objectives, the trajectory sources, and the supervision used for each update.

}

\bibliography{current_references}
\bibliographystyle{iclr2027_conference}
\clearpage
\appendix
\onecolumn

\section{Environments and Rewards}
\label{appendix:datasets}
We describe the task objectives, reward signals, and demonstration sources for the three environments. All experiments use episode-level rewards in $[0,1]$. {We use a maximum interaction budget of 10 turns per episode for WebShop and InterCodeSQL and 100 turns for ScienceWorld.} Dataset sizes and interaction budgets are summarized in Table~\ref{tab:overview}.

\subsection{WebShop}
WebShop requires an agent to select a product and options that satisfy a shopping instruction~\citep{webshop}. Its terminal reward combines matches with the requested attributes and options, satisfaction of the price constraint, and a match with the requested product type. Partial matches receive partial credit. We use the demonstration trajectories provided by \citet{eto}.

\subsection{ScienceWorld}
ScienceWorld requires navigation, object manipulation, and scientific procedures~\citep{scienceworld}. We normalize its task progress score to $[0,1]$. The score reflects completion of intermediate goals, so unsuccessful trajectories can receive partial credit. Following the evaluation setup used in our experiments, we exclude the two longest task types and subsample test instances with more than ten variations. Seen scenarios and unseen variations are evaluated separately. Demonstrations are obtained from the official data and GPT-4-generated trajectories following \citet{eto}.

\subsection{InterCodeSQL}
InterCodeSQL requires an agent to answer a database question through interactive SQL execution~\citep{intercode}. The reward compares the final execution output with the reference output using an intersection-over-union score that accounts for matching records, duplicates, and record order. Invalid execution outputs receive zero reward, partial matches receive intermediate rewards, and an exact correctly ordered match receives one. We use GPT-4 demonstrations collected by \citet{IPR}, retaining those with reward $1.0$, and add ReAct-style thought annotations following that setup~\citep{react}.

\section{Training Details}
\label{appendix:Implementation}
\subsection{Optimization}
Both agents are initialized from the same supervised checkpoint. The recorded SFT configuration uses AdamW for three epochs with batch size $48$, learning rate $10^{-5}$, a $3\%$ linear warmup, and cosine decay. Preference optimization uses batch size $32$ and three epochs per phase, with learning rates between $5\times10^{-7}$ and $10^{-6}$ and DPO scale $\beta$ between $0.1$ and $0.5$. Experiments use eight NVIDIA H100 GPUs with 80 GB memory each. The main target objective uses $\lambda_{\mathrm{SFT}}=0.1$.

\subsection{Preference Data and Sample Weighting}
\label{appendix:weighting}
The target agent is trained on expert--failure pairs: the expert demonstration is preferred, and the paired failure comes from either agent. The auxiliary SFT term uses the expert demonstration from the same pair. The failure agent uses only failure--failure preferences and has no auxiliary SFT term after initialization.

To give each instruction equal total weight, let $m_u>0$ denote the number of retained target preference pairs for instruction $u$. Starting from uniform pair weights, we assign each pair weight $1/m_u$. If $w_i$ is the resulting weight of pair $i$, the empirical average in Equation~\ref{eq:target_objective} is
\begin{equation}
\mathbb{E}_w[f]=\frac{\sum_i w_i f_i}{\sum_i w_i}.
\label{eq:appendix_weighting}
\end{equation}
This normalization prevents instructions with more collected failures from contributing more total weight. It is applied to both the target DPO term and its auxiliary SFT term.

\FloatBarrier
\section{Additional Experiments}
\label{appendix:additional_results}
\subsection{Additional Model Families}
We report additional comparisons using the earlier Llama-2-7B-Chat and Llama-2-13B-Chat models in Tables~\ref{tab:main} and~\ref{tab:llama2_13b}. Our framework exceeds ETO in every reported task column for both models. Each table reports a separate experimental configuration; the results should be compared within the corresponding setting. GPT results in Table~\ref{tab:main} use in-context learning rather than task-specific fine-tuning.

\begin{table}[!htbp]
\color{black}
\centering
\caption{Additional results for Llama-2-7B-Chat. GPT baselines use in-context learning. Avg. is the mean over the four evaluation settings.}
\label{tab:main}
\small
\begin{tabular}{clccccc}
\toprule
\multirow{2}{*}{\textbf{Adaptation}} &
\multirow{2}{*}{\textbf{Models}} &
\multirow{2}{*}{\textbf{WebShop}} &
\multicolumn{2}{c}{\textbf{ScienceWorld}} &
\multirow{2}{*}{\textbf{InterCodeSQL}} &
\multirow{2}{*}{\textbf{Avg.}} \\
\cmidrule(lr){4-5}
& & & \textbf{Seen} & \textbf{Unseen} & & \\
\midrule
\multirow{2}{*}{In-context}
& GPT-4 & 63.2 & 42.9 & 38.1 & 38.5 & 45.7 \\
& GPT-3.5-Turbo & 62.4 &  7.9 & 10.5 & 37.8 & 29.7 \\
\midrule
\multirow{5}{*}{Fine-tuning}
& Llama-2-7B-Chat + SFT & 58.7 & 53.4 & 49.9 & 26.5 & 47.1 \\
& Llama-2-7B-Chat + PPO & 64.2 & 59.4 & 51.7 & 32.4 & 51.9 \\
& Llama-2-7B-Chat + RFT & 61.3 & 71.6 & 54.3 & 25.6 & 53.2 \\
& Llama-2-7B-Chat + ETO & 63.0 & 73.1 & 65.3 & 32.3 & 58.4 \\
& Llama-2-7B-Chat + Ours & \textbf{70.3} & \textbf{82.4} & \textbf{69.2} & \textbf{43.4} & \textbf{66.3} \\
\bottomrule
\end{tabular}

\end{table}

\begin{table}[!htbp]
\color{black}
\centering
\caption{Additional results for Llama-2-13B-Chat on WebShop and ScienceWorld. Avg. is the mean over the two environments.}
\label{tab:llama2_13b}
\small
\setlength{\tabcolsep}{3pt}
\begin{tabular*}{\linewidth}{@{\extracolsep{\fill}}lccc}
\toprule
\textbf{Models} &
\textbf{WebShop} &
\textbf{ScienceWorld} &
\textbf{Avg.} \\
\midrule
Llama-2-13B-chat + SFT        & 61.8 & 64.5 & 63.2 \\
Llama-2-13B-chat + AgentLM~\citep{agenttuning}    & 39.5 & 10.7 & 25.1 \\
Llama-2-13B-chat + Agent-FLAN~\citep{agentflan} & 40.4 & 28.6 & 34.5 \\
Llama-2-13B-chat + ETO        & 66.0 & 72.6 & 69.3 \\
Llama-2-13B-chat + Ours       & \textbf{69.2} & \textbf{74.5} & \textbf{71.9} \\
\bottomrule
\end{tabular*}
\end{table}

\FloatBarrier
\subsection{Additional Sampling for RFT}
We examine whether allocating more samples to difficult instructions improves RFT, which learns from successful trajectories. A DART-style procedure uses five initial rollouts to estimate instruction difficulty and allocates a total of ten candidates per instruction. Successful candidates augment the supervised training data. In this separate WebShop experiment, the score increases from $60.89$ to $61.90$, a gain of $1.01$ points. The result supports additional sampling for collecting successful training trajectories in this setting. It addresses a different question from Best-of-$N$, which selects trajectories for preference learning, and uses a separate configuration from the RFT results in the main table.

\subsection{Sensitivity to the SFT Weight}
\label{appendix:sft_weight}
Table~\ref{tab:lambda_ablation_llama2} reports sensitivity to the SFT weight for Llama-2-7B on WebShop. The highest score occurs at $\lambda_{\mathrm{SFT}}=0.01$; increasing the weight does not improve performance over the tested range.

\begin{center}
\captionsetup{hypcap=false}
\captionof{table}{Effect of the SFT weight on Llama-2-7B performance in WebShop.}
\label{tab:lambda_ablation_llama2}
\small
\setlength{\tabcolsep}{10pt}
\begin{tabular}{cc}
\toprule
$\lambda_{\mathrm{SFT}}$ & WebShop \\
\midrule
0.01 & 67.98 \\
0.1 & 66.70 \\
0.5 & 66.50 \\
\bottomrule
\end{tabular}
\end{center}

\FloatBarrier
\section{Qualitative Results for Qwen3.5 and Gemma-4}
\label{appendix:quality}
We present three matched comparisons per environment: two target evaluations and one comparison of training failures. Target evaluations compare ETO with our target agent; training comparisons contrast failures from ETO's target agent with those from our failure agent. Each case reports the model, task ID, recorded reward, and outcome. Rewards are on the original $[0,1]$ scale. We summarize the behavior and show selected actions and environment responses. ScienceWorld target examples use the unseen split. Evaluation trajectories are from \texttt{data/qualitative\_trajs}; training failures are from \texttt{data/train\_failures}.

\subsection{ScienceWorld}
\label{appendix:qual_sciworld}
\begin{tcolorbox}[trajbox,lines before break=8,before skip=5pt,after skip=5pt,boxsep=1pt,top=3pt,bottom=3pt,title={Completing the Pollination Step},title after break={Completing the Pollination Step (continued)}]
\textbf{Setting:} Gemma-4-12B-it; task \texttt{15\_93}. Target evaluation.\par\smallskip
\textbf{Instruction:} Grow an orange by growing several plants and allowing cross-pollination. Focus on the grown orange to finish the task.\par\vspace{3pt}\hrule\vspace{3pt}
\textbf{ETO target} \hfill Reward 0.446 \quad \qualoutcome{qualFailure}{FAILURE}\par\nobreak
Plants three orange seeds and waits until the trees have flowers, but \textbf{leaves the bee hive closed}. It continues waiting and attempts to focus on an orange before any fruit appears, leaving the pollination step incomplete.
\par\nobreak\smallskip
{\small\raggedright Environment: ``The bee hive door is closed.'' Later attempts to \texttt{focus on orange in reproducing orange tree} receive ``No known action matches that input.''\par}
\par\vspace{3pt}\hrule\vspace{3pt}
\textbf{Ours target} \hfill Reward 1.000 \quad \qualoutcome{qualSuccess}{SUCCESS}\par\nobreak
Plants the seeds, waits for flowers, and \textbf{opens the bee hive}. Later observations show a bee and oranges on the trees; after resolving an ambiguous object selection, it \textbf{focuses on a grown orange and completes the task}.
\par\nobreak\smallskip
{\small\raggedright Action: \texttt{open bee hive}. Later observations list an adult bee and oranges; \texttt{focus on orange} followed by object selection yields ``You focus on the orange.''\par}
\par\nobreak\vspace{3pt}\hrule\nobreak\vspace{3pt}
\textbf{Key difference:} Both agents grow flowering trees, but only Ours enables pollination and follows the resulting fruit through to task completion.
\end{tcolorbox}

\begin{tcolorbox}[trajbox,lines before break=8,before skip=5pt,after skip=5pt,boxsep=1pt,top=3pt,bottom=3pt,title={Using Feedback to Complete a Circuit},title after break={Using Feedback to Complete a Circuit (continued)}]
\textbf{Setting:} Qwen3.5-9B; task \texttt{10\_451}. Target evaluation.\par\smallskip
\textbf{Instruction:} Test whether unknown substance U conducts electricity. Place it in the green box if conductive, or the blue box otherwise.\par\vspace{3pt}\hrule\vspace{3pt}
\textbf{ETO target} \hfill Reward 0.643 \quad \qualoutcome{qualFailure}{FAILURE}\par\nobreak
Builds part of a battery-and-bulb circuit, then repeatedly tries to connect wires to \textbf{an anode or cathode on the substance}. The environment rejects these actions; repeated waiting does not repair the connection, and the material is not classified.
\par\nobreak\smallskip
{\small\raggedright Action: \texttt{connect blue wire terminal 2 to cathode in unknown substance U}. Response: ``No known action matches that input.''\par}
\par\vspace{3pt}\hrule\vspace{3pt}
\textbf{Ours target} \hfill Reward 1.000 \quad \qualoutcome{qualSuccess}{SUCCESS}\par\nobreak
Connects the substance through its \textbf{numbered terminals} and completes the circuit. After waiting, it observes that \textbf{the red bulb is on} and moves the substance to the green box, linking the observed electrical response to the required decision.
\par\nobreak\smallskip
{\small\raggedright Connections use \texttt{unknown substance U terminal 1} and \texttt{terminal 2}. The room description changes to ``a red light bulb, which is on''; the environment then confirms placement in the green box.\par}
\par\nobreak\vspace{3pt}\hrule\nobreak\vspace{3pt}
\textbf{Key difference:} Ours completes the experimental setup and uses its observed outcome to classify the material, whereas ETO repeats an invalid connection.
\end{tcolorbox}

\begin{tcolorbox}[trajbox,lines before break=8,before skip=5pt,after skip=5pt,boxsep=1pt,top=3pt,bottom=3pt,title={A Measurement without a Valid Conclusion},title after break={A Measurement without a Valid Conclusion (continued)}]
\textbf{Setting:} Gemma-4-12B-it; task \texttt{4\_22}. Training failures, iteration 2.\par\smallskip
\textbf{Instruction:} Measure the melting point of gallium. Focus on the orange box if it is above $200^\circ$C, or the yellow box if it is below $200^\circ$C.\par\vspace{3pt}\hrule\vspace{3pt}
\textbf{ETO negative} \hfill Reward 0.403 \quad \qualoutcome{qualFailure}{FAILURE}\par\nobreak
Retrieves and focuses on the thermometer, then attempts to focus on the gallium-containing jug. The environment \textbf{selects the jug itself}, and the rollout \textbf{ends before the agent measures the material} or selects an answer box.
\par\nobreak\smallskip
{\small\raggedright Action: \texttt{focus on jug containing gallium}. Response: ``You focus on the jug.'' No thermometer reading follows.\par}
\par\vspace{3pt}\hrule\vspace{3pt}
\textbf{Ours hard negative} \hfill Reward 0.747 \quad \qualoutcome{qualFailure}{FAILURE}\par\nobreak
Focuses on gallium, opens the freezer, cools the sample, and obtains a $7^\circ$C reading. It nevertheless \textbf{selects the orange box for the above-$200^\circ$C condition}; the measurement also \textbf{does not establish the melting point} requested by the instruction.
\par\nobreak\smallskip
{\small\raggedright Action: \texttt{use thermometer in inventory on gallium}. Response: ``the thermometer measures a temperature of 7 degrees celsius''. The next action is \texttt{focus on orange box}.\par}
\par\nobreak\vspace{3pt}\hrule\nobreak\vspace{3pt}
\textbf{Key difference:} The hard negative completes more of the procedure but still reaches an unsupported conclusion. The error concerns what the measurement establishes and how it should guide the final decision.
\end{tcolorbox}

\subsection{WebShop}
\label{appendix:qual_webshop}
\begin{tcolorbox}[trajbox,lines before break=8,before skip=5pt,after skip=5pt,boxsep=1pt,top=3pt,bottom=3pt,title={Preserving the Requested Size},title after break={Preserving the Requested Size (continued)}]
\textbf{Setting:} Gemma-4-12B-it; task \texttt{11151}. Target evaluation.\par\smallskip
\textbf{Instruction:} Buy men's size-14 sneakers with a synthetic sole for less than \$140.\par\vspace{3pt}\hrule\vspace{3pt}
\textbf{ETO target} \hfill Reward 0.667 \quad \qualoutcome{qualFailure}{FAILURE}\par\nobreak
Opens a Madden sneaker listing whose available sizes stop at 13. It \textbf{selects size 13} and buys the product, completing the purchase flow but dropping the explicit size requirement from the instruction.
\par\nobreak\smallskip
{\small\raggedright Product: \texttt{B08L9SFY3C}. Actions include \texttt{click[13]} and \texttt{click[Buy Now]}.\par}
\par\vspace{3pt}\hrule\vspace{3pt}
\textbf{Ours target} \hfill Reward 1.000 \quad \qualoutcome{qualSuccess}{SUCCESS}\par\nobreak
Selects an ALDO sneaker listing that \textbf{offers size 14, chooses that size}, and completes the purchase. The recorded actions preserve the requested size through product selection and checkout rather than substituting an available smaller option.
\par\nobreak\smallskip
{\small\raggedright Product: \texttt{B08SPLLMN3}. Actions include \texttt{click[14]} and \texttt{click[Buy Now]}.\par}
\par\nobreak\vspace{3pt}\hrule\nobreak\vspace{3pt}
\textbf{Key difference:} The distinction is the selected product option: completing a purchase does not suffice when the requested size is unavailable.
\end{tcolorbox}

\begin{tcolorbox}[trajbox,lines before break=8,before skip=5pt,after skip=5pt,boxsep=1pt,top=3pt,bottom=3pt,title={Checking the Budget before Purchase},title after break={Checking the Budget before Purchase (continued)}]
\textbf{Setting:} Qwen3.5-9B; task \texttt{1414}. Target evaluation.\par\smallskip
\textbf{Instruction:} Buy green, modern velvet dining chairs for less than \$210.\par\vspace{3pt}\hrule\vspace{3pt}
\textbf{ETO target} \hfill Reward 0.667 \quad \qualoutcome{qualFailure}{FAILURE}\par\nobreak
Finds green velvet dining chairs, selects the green option, and \textbf{buys a set listed at \$260.99}. The product matches the requested style and color, but its displayed price exceeds the budget before checkout.
\par\nobreak\smallskip
{\small\raggedright Product: \texttt{B07YTMKL9T}; displayed price: \$260.99. Actions: \texttt{click[green]}, then \texttt{click[Buy Now]}.\par}
\par\vspace{3pt}\hrule\vspace{3pt}
\textbf{Ours target} \hfill Reward 1.000 \quad \qualoutcome{qualSuccess}{SUCCESS}\par\nobreak
Chooses a different velvet dining-chair listing \textbf{priced at \$173.99}, selects emerald green, and completes the purchase. This keeps the requested appearance while satisfying the price constraint, rather than accepting the first visually relevant product.
\par\nobreak\smallskip
{\small\raggedright Product: \texttt{B07R9KFCKY}; displayed price: \$173.99. Actions: \texttt{click[emerald green]}, then \texttt{click[Buy Now]}.\par}
\par\nobreak\vspace{3pt}\hrule\nobreak\vspace{3pt}
\textbf{Key difference:} Both agents choose relevant chairs; the successful trajectory also respects the budget at the purchase decision.
\end{tcolorbox}

\begin{tcolorbox}[trajbox,lines before break=8,before skip=5pt,after skip=5pt,boxsep=1pt,top=3pt,bottom=3pt,title={Relevant Shoes with a Missed Budget Constraint},title after break={Relevant Shoes with a Missed Budget Constraint (continued)}]
\textbf{Setting:} Gemma-4-12B-it; task \texttt{7092}. Training failures, iteration 1.\par\smallskip
\textbf{Instruction:} Buy breathable, slip-resistant men's shoes with rubber soles, size 8, for less than \$80.\par\vspace{3pt}\hrule\vspace{3pt}
\textbf{ETO negative} \hfill Reward 0.075 \quad \qualoutcome{qualFailure}{FAILURE}\par\nobreak
\textbf{Selects a hiking sneaker} listed at \$42.88, chooses size 8 and black, and completes the purchase. Although the recorded size and price satisfy those two requirements, the overall product match receives only a small partial reward.
\par\nobreak\smallskip
{\small\raggedright Product: \texttt{B07S7QXNCW}; price: \$42.88. Actions include \texttt{click[8]}, \texttt{click[black]}, and \texttt{click[Buy Now]}.\par}
\par\vspace{3pt}\hrule\vspace{3pt}
\textbf{Ours hard negative} \hfill Reward 0.750 \quad \qualoutcome{qualFailure}{FAILURE}\par\nobreak
Chooses shoes described as \textbf{breathable and non-slip, selects size 8}, and buys them. The listing is \textbf{priced at \$100}, so the trajectory preserves several relevant choices but \textbf{fails to enforce the explicit budget constraint} at checkout.
\par\nobreak\smallskip
{\small\raggedright Product: \texttt{B09CYSLPZL}; title includes ``Breathable Non-Slip''; price: \$100. Actions include \texttt{click[8]} and \texttt{click[Buy Now]}.\par}
\par\nobreak\vspace{3pt}\hrule\nobreak\vspace{3pt}
\textbf{Key difference:} The higher-reward failure is a plausible purchase with a specific unmet constraint. Comparing it with a successful demonstration can expose the missed price check.
\end{tcolorbox}

\subsection{InterCodeSQL}
\label{appendix:qual_intercode_sql}
\begin{tcolorbox}[trajbox,lines before break=8,before skip=5pt,after skip=5pt,boxsep=1pt,top=3pt,bottom=3pt,title={Returning All Requested Fields},title after break={Returning All Requested Fields (continued)}]
\textbf{Setting:} Gemma-4-12B-it; task \texttt{82}. Target evaluation.\par\smallskip
\textbf{Instruction:} Find the language used by the fewest TV channels, and return both the language and the number of channels.\par\vspace{3pt}\hrule\vspace{3pt}
\textbf{ETO target} \hfill Reward 0.000 \quad \qualoutcome{qualFailure}{FAILURE}\par\nobreak
Groups channels by language and orders the groups by their counts, but \textbf{selects only the language column}. The query executes and returns English; the agent submits this incomplete output without the requested channel count.
\par\nobreak\smallskip
{\small\raggedright Query fragment: \texttt{SELECT Language FROM TV\_Channel GROUP BY Language ORDER BY count(*) LIMIT 1}. Output: \texttt{[(\textquotesingle English\textquotesingle,)]}.\par}
\par\vspace{3pt}\hrule\vspace{3pt}
\textbf{Ours target} \hfill Reward 1.000 \quad \qualoutcome{qualSuccess}{SUCCESS}\par\nobreak
Initially obtains the same language-only output, then \textbf{revises the projection to include the count}. It submits the resulting language--count pair, correcting the missing output field while retaining the grouping and ordering needed by the question.
\par\nobreak\smallskip
{\small\raggedright Query fragment: \texttt{SELECT Language, COUNT(*) FROM TV\_Channel}. The final output is \texttt{[(\textquotesingle English\textquotesingle, 3)]}.\par}
\par\nobreak\vspace{3pt}\hrule\nobreak\vspace{3pt}
\textbf{Key difference:} Successful execution alone does not establish task success: the returned fields must also match the question.
\end{tcolorbox}

\begin{tcolorbox}[trajbox,lines before break=8,before skip=5pt,after skip=5pt,boxsep=1pt,top=3pt,bottom=3pt,title={Using the Stored Categorical Value},title after break={Using the Stored Categorical Value (continued)}]
\textbf{Setting:} Gemma-4-12B-it; task \texttt{114}. Target evaluation.\par\smallskip
\textbf{Instruction:} How many dog pets are raised by female students?\par\vspace{3pt}\hrule\vspace{3pt}
\textbf{ETO target} \hfill Reward 0.000 \quad \qualoutcome{qualFailure}{FAILURE}\par\nobreak
Repairs initial query errors and joins the student, ownership, and pet tables. Its \textbf{final filter uses \texttt{Sex = \textquotesingle female\textquotesingle}}; this query executes but returns zero, and the agent submits that result without correcting the stored category value.
\par\nobreak\smallskip
{\small\raggedright Final predicates: \texttt{student.Sex = \textquotesingle female\textquotesingle} and \texttt{pets.PetType = \textquotesingle dog\textquotesingle}. Output: \texttt{[(0,)]}.\par}
\par\vspace{3pt}\hrule\vspace{3pt}
\textbf{Ours target} \hfill Reward 1.000 \quad \qualoutcome{qualSuccess}{SUCCESS}\par\nobreak
Inspects the table schemas and joins the same entities using their identifiers. It \textbf{filters students with \texttt{Sex = \textquotesingle F\textquotesingle}} and pets with \texttt{PetType = \textquotesingle dog\textquotesingle}, obtaining and submitting a count of two.
\par\nobreak\smallskip
{\small\raggedright The schema gives \texttt{Sex} as \texttt{varchar(1)}. The final query uses \texttt{Sex = \textquotesingle F\textquotesingle} and returns \texttt{[(2,)]}.\par}
\par\nobreak\vspace{3pt}\hrule\nobreak\vspace{3pt}
\textbf{Key difference:} A valid join can still yield an incorrect answer when the filter uses a natural-language label instead of the value stored in the database.
\end{tcolorbox}

\begin{tcolorbox}[trajbox,lines before break=8,before skip=5pt,after skip=5pt,boxsep=1pt,top=3pt,bottom=3pt,title={A Plausible Exclusion Query with the Wrong Identifier},title after break={A Plausible Exclusion Query with the Wrong Identifier (continued)}]
\textbf{Setting:} Gemma-4-12B-it; task \texttt{2351}. Training failures, iteration 2.\par\smallskip
\textbf{Instruction:} Return the titles and directors of films that were never presented in China.\par\vspace{3pt}\hrule\vspace{3pt}
\textbf{ETO negative} \hfill Reward 0.000 \quad \qualoutcome{qualFailure}{FAILURE}\par\nobreak
Inspects the schemas after two invalid queries, then filters joined records for China while \textbf{requiring the film title to be null}. This does not express the requested exclusion, and the submitted query returns an empty result.
\par\nobreak\smallskip
{\small\raggedright Final predicates include \texttt{T3.country = \textquotesingle China\textquotesingle AND T1.title IS NULL}. Output: \texttt{[]}.\par}
\par\vspace{3pt}\hrule\vspace{3pt}
\textbf{Ours hard negative} \hfill Reward 0.770 \quad \qualoutcome{qualFailure}{FAILURE}\par\nobreak
Uses a plausible \texttt{NOT IN} query and returns film titles and directors, earning substantial partial credit. However, it \textbf{compares the numeric market identifier with \texttt{\textquotesingle CH\textquotesingle}} instead of resolving China through the market table, so the exclusion condition is incorrect.
\par\nobreak\smallskip
{\small\raggedright Exclusion subquery: \texttt{SELECT film\_id FROM film\_market\_estimation WHERE market\_id = \textquotesingle CH\textquotesingle}. The schema identifies \texttt{Market\_ID} as an integer; the submitted output contains 13 title--director pairs.\par}
\par\nobreak\vspace{3pt}\hrule\nobreak\vspace{3pt}
\textbf{Key difference:} The hard negative produces a nonempty, partially correct answer with the expected columns, but a mistaken identifier invalidates the condition that determines which rows belong in the result.
\end{tcolorbox}

\FloatBarrier
\section{Qualitative Results for Llama-2}
\label{appendix:llama2_qualitative}
\label{sec:llama2_qualitative}
\label{appendix:llama2_quality}
This section presents qualitative examples from the Llama-2 experiments, complementing the Qwen3.5 and Gemma-4 comparisons in Appendix~\ref{appendix:quality}. We preserve the reported rewards and step counts, but do not infer exact failure causes or answer correctness from partial reward alone.

\subsection{WebShop}
\begin{tcolorbox}[trajbox,title={HDMI Cables under \$50},title after break={HDMI Cables under \$50 (continued)}]
\textbf{Instruction:}
I'm looking for ten high-speed, gold-plated HDMI cables, with price lower than \$50.00.

\medskip
\hrule
\medskip

\textbf{ETO:}
Selects a single ProHT HDMI cable priced at \$100 and proceeds to purchase. The selected quantity and price do not match the instruction.
\par\nobreak
\emph{Reward:} 0.50 \quad
\emph{Steps:} 4 \quad
\emph{Outcome:} Failure
\medskip
\hrule
\medskip

\textbf{Ours:}
Searches for multi-pack HDMI cables, inspects a QualGear candidate, and selects a variant. The summary reports checks of cable properties and price, but the exact pack quantity remains unresolved.
\par\nobreak
\emph{Reward:} 0.75 \quad
\emph{Steps:} 5 \quad
\emph{Outcome:} Failure

\medskip
\hrule
\medskip

\textbf{Interpretation:}
The higher-reward failure includes more directed product inspection.
\end{tcolorbox}

\begin{tcolorbox}[trajbox,title={Solid Wood Storage Bench in Grey},title after break={Solid Wood Storage Bench in Grey (continued)}]
\textbf{Instruction:}
I want a solid wood bench with storage space for my living room, grey in color, and under \$210.00.

\medskip
\hrule
\medskip

\textbf{ETO:}
Selects a grey accent bench after a short search and proceeds to purchase.
\par\nobreak
\emph{Reward:} 0.50 \quad
\emph{Steps:} 3 \quad
\emph{Outcome:} Failure
\medskip
\hrule
\medskip

\textbf{Ours:}
Inspects several candidates and selects a grey storage bench, with reasoning about its material, storage features, and price.
\par\nobreak
\emph{Reward:} 0.75 \quad
\emph{Steps:} 5 \quad
\emph{Outcome:} Failure

\medskip
\hrule
\medskip

\textbf{Interpretation:}
The traces differ in product inspection and selection. The reported reward does not verify all product attributes or establish that price is the sole unmet requirement.
\end{tcolorbox}

\begin{tcolorbox}[trajbox,title={Machine-Washable Curtains (52"×90")},title after break={Machine-Washable Curtains (52"×90") (continued)}]

\textbf{Instruction:}
I need a machine-washable curtain for the living room, sized 52" wide by 90" long, priced under \$60.00.

\medskip
\hrule
\medskip

\textbf{ETO:}
Selects an early search result and the 52" by 90" option, then attempts a purchase.
\par\nobreak
\emph{Reward:} 0.50 \quad
\emph{Steps:} 4 \quad
\emph{Outcome:} Failure

\medskip
\hrule
\medskip
\textbf{Ours:}
Visits multiple product pages, reasons about washability and price, selects the requested size, and attempts a purchase.
\par\nobreak
\emph{Reward:} 0.75 \quad
\emph{Steps:} 8 \quad
\emph{Outcome:} Failure

\medskip
\hrule
\medskip

\textbf{Interpretation:}
Both trajectories remain unsuccessful. Additional attribute checks are present in the higher-reward summary, but they do not establish that every constraint was satisfied.
\end{tcolorbox}

\subsection{ScienceWorld}
\begin{tcolorbox}[trajbox,title={Moving a Non-Living Object to the Green Box},title after break={Moving a Non-Living Object to the Green Box (continued)}]

\textbf{Instruction:}
Find a non-living object, focus on it, and move it to the green box in the workshop.

\medskip
\hrule
\medskip

\textbf{ETO:}
Moves a yellow wire into the green box, then repeats waiting and inspection actions without an explicit focus step.
\par\nobreak
\emph{Reward:} 0.25 \quad
\emph{Steps:} 15 \quad
\emph{Outcome:} Failure

\medskip
\hrule
\medskip
\textbf{Ours:}
Moves the same wire into the green box and issues focus actions on the box and the wire.
\par\nobreak
\emph{Reward:} 0.75 \quad
\emph{Steps:} 15 \quad
\emph{Outcome:} Failure

\medskip
\hrule
\medskip

\textbf{Interpretation:}
The explicit focus actions distinguish the two attempts. The higher reward does not establish that all required actions were accepted or that the environment incorrectly withheld success.
\end{tcolorbox}

\begin{tcolorbox}[trajbox,title={Turning On a Green Light Bulb with Renewable Power},title after break={Turning On a Green Light Bulb with Renewable Power (continued)}]
\textbf{Instruction:}
Your task is to turn on the green light bulb. First, focus on the green light bulb. Then, create an electrical circuit that powers it on. Prefer renewable power sources when possible.

\medskip
\hrule
\medskip

\textbf{ETO:}
Explores circuit components and attempts connections, interspersed with focus actions on other objects.
\par\nobreak
\emph{Reward:} 0.43 \quad \emph{Steps:} 30 \quad \emph{Outcome:} Failure

\medskip
\hrule
\medskip
\textbf{Ours:}
Focuses on the green bulb and connects wires involving the bulb and a renewable power component, repeatedly checking the state.
\par\nobreak
\emph{Reward:} 0.58 \quad \emph{Steps:} 30 \quad \emph{Outcome:} Failure

\medskip
\hrule
\medskip

\textbf{Interpretation:}
The higher-reward summary contains a more task-directed sequence.
\end{tcolorbox}

\begin{tcolorbox}[trajbox,title={Growing a Banana from Seed to Fruit},title after break={Growing a Banana from Seed to Fruit (continued)}]

\textbf{Instruction:}
Your task is to grow a banana. This requires obtaining banana seeds, planting them in soil, providing water and light, and waiting until the banana grows.

\medskip
\hrule
\medskip

\textbf{ETO:}
Collects seeds and containers, with repeated navigation and inspection actions.
\par\nobreak
\emph{Reward:} 0.36 \quad \emph{Steps:} 55 \quad \emph{Outcome:} Failure

\medskip
\hrule
\medskip
\textbf{Ours:}
Collects banana seeds and performs planting, watering, and growth-monitoring actions.
\par\nobreak
\emph{Reward:} 0.50 \quad \emph{Steps:} 60 \quad \emph{Outcome:} Failure

\medskip
\hrule
\medskip

\textbf{Interpretation:}
The higher-reward failure includes several stages of the intended procedure. It does not establish that all growth conditions were met.
\end{tcolorbox}

\subsection{InterCodeSQL}
\begin{tcolorbox}[trajbox,title={Films Not Presented in China},title after break={Films Not Presented in China (continued)}]

\textbf{Instruction:}
List the titles and directors of films that were never presented in China.

\medskip
\hrule
\medskip

\textbf{ETO:}
Inspects tables and repeatedly queries fields such as \texttt{Market} and \texttt{country} while attempting joins.
\par\nobreak
\emph{Reward:} 0.00 \quad
\emph{Steps:} 6 \quad
\emph{Outcome:} Failure

\medskip
\hrule
\medskip
\textbf{Ours:}
Inspects schemas, revises table and column choices, and constructs a query through \texttt{market} for film titles and directors.
\par\nobreak
\emph{Reward:} 0.77 \quad
\emph{Steps:} 9 \quad
\emph{Outcome:} Failure

\medskip
\hrule
\medskip

\textbf{Interpretation:}
The summary shows a revised query structure and higher partial reward. Executability and exact answer correctness must be distinguished; the reported result remains a failure.
\end{tcolorbox}

\begin{tcolorbox}[trajbox,title={Reviewers Who Rated Above 3 Stars},title after break={Reviewers Who Rated Above 3 Stars (continued)}]

\textbf{Instruction:}
Find the names of reviewers who previously rated a movie more than 3 stars.

\medskip
\hrule
\medskip

\textbf{ETO:}
Attempts joins across \texttt{reviewer}, \texttt{rating}, and \texttt{movie}, repeatedly revisiting their schemas.
\par\nobreak
\emph{Reward:} 0.00 \quad
\emph{Steps:} 10 \quad
\emph{Outcome:} Failure

\medskip
\hrule
\medskip
\textbf{Ours:}
Inspects reviewer and rating fields and constructs a join using \texttt{rID} to retrieve reviewer names.
\par\nobreak
\emph{Reward:} 0.75 \quad
\emph{Steps:} 5 \quad
\emph{Outcome:} Failure

\medskip
\hrule
\medskip

\textbf{Interpretation:}
The revised join receives partial reward. The record does not justify calling the answer fully correct when the reported outcome is failure.
\end{tcolorbox}

\begin{tcolorbox}[trajbox,title={Gymnasts Ordered by Ascending Height},title after break={Gymnasts Ordered by Ascending Height (continued)}]

\textbf{Instruction:}
Return the names of gymnasts ordered by their height in ascending order.

\medskip
\hrule
\medskip

\textbf{ETO:}
Attempts to retrieve names and heights from \texttt{gymnast}, with repeated schema inspection.
\par\nobreak
\emph{Reward:} 0.00 \quad
\emph{Steps:} 6 \quad
\emph{Outcome:} Failure

\medskip
\hrule
\medskip
\textbf{Ours:}
Inspects both \texttt{gymnast} and \texttt{people}, then uses the name and height fields in a height-ordered query.
\par\nobreak
\emph{Reward:} 0.70 \quad
\emph{Steps:} 6 \quad
\emph{Outcome:} Failure

\medskip
\hrule
\medskip

\textbf{Interpretation:}
The summary shows a change in table selection and ordering.
\end{tcolorbox}

\subsection{Training Failures and Target Behavior}
\label{appendix:llama2_effect}
The lemon-growing example compares unsuccessful training trajectories with subsequent target rollouts. The trajectory produced by the failure agent includes soil preparation and planting, while the baseline trajectory repeats a narrower set of actions. The target rollout produced by our method succeeds. The example documents this difference but does not show that a specific failure caused the successful target rollout.

\begin{tcolorbox}[trajbox,title={Growing a Lemon with Cross-Pollination},title after break={Growing a Lemon with Cross-Pollination (continued)}]

\textbf{Instruction:}
Your task is to grow a lemon. This will require growing several plants and having them cross-pollinated to produce fruit. 
Seeds can be found in the bedroom. To complete the task, focus on the grown lemon.

\medskip
\hrule
\medskip

\textbf{ETO: Target Rollout}
Retrieves the seed jar, plants seeds in water-filled pots, and repeats waiting, inspection, and focus attempts. The summary records no soil preparation.
\par\nobreak
\emph{Reward:} 0.25 \quad
\emph{Steps:} 60 \quad
\emph{Outcome:} Failure

\textbf{ETO: Training Failure}
Plants seeds in water-filled pots and repeatedly checks the greenhouse and attempts to focus on the lemon tree.
\par\nobreak
\emph{Reward:} 0.25 \quad
\emph{Steps:} 49 \quad
\emph{Outcome:} Failure

\medskip
\hrule
\medskip

\textbf{Ours: Target Rollout}
Collects soil, fills the pots, plants seeds, waits for growth, closes greenhouse doors, and focuses on the lemon. The reported outcome is success.
\par\nobreak
\emph{Reward:} 1.00 \quad
\emph{Steps:} 46 \quad
\emph{Outcome:} Success

\textbf{Ours: Training Failure}
Collects and transports soil, fills pots, plants seeds, and later attempts focus and pickup actions referring to the lemon. The reported outcome remains failure.
\par\nobreak
\emph{Reward:} 0.50 \quad
\emph{Steps:} 60 \quad
\emph{Outcome:} Failure

\medskip
\hrule
\end{tcolorbox}

\FloatBarrier

\end{document}